\documentclass{article}

\usepackage[final,nonatbib]{nips_2016}

\usepackage[utf8]{inputenc} 
\usepackage[T1]{fontenc}    
\usepackage{hyperref}       
\usepackage{url}            
\usepackage{booktabs}       
\usepackage{amsfonts}       
\usepackage{nicefrac}       
\usepackage{microtype}      
\usepackage{amsmath}
\usepackage{graphicx}
\usepackage{subcaption}

\usepackage{xcolor}
\usepackage{listings}

\title{An overview of gradient descent optimization algorithms\thanks{This paper originally appeared as a blog post at \url{http://sebastianruder.com/optimizing-gradient-descent/index.html} on 19 January 2016.}}

\author{
  Sebastian Ruder\\
  Insight Centre for Data Analytics, NUI Galway\\
  Aylien Ltd., Dublin\\
  \texttt{ruder.sebastian@gmail.com}
}

\begin{document}

\maketitle

\begin{abstract}
Gradient descent optimization algorithms, while increasingly popular, are often used as black-box optimizers, as practical explanations of their strengths and weaknesses are hard to come by. This article aims to provide the reader with intuitions with regard to the behaviour of different algorithms that will allow her to put them to use. In the course of this overview, we look at different variants of gradient descent, summarize challenges, introduce the most common optimization algorithms, review architectures in a parallel and distributed setting, and investigate additional strategies for optimizing gradient descent.
\end{abstract}

\section{Introduction}

Gradient descent is one of the most popular algorithms to perform optimization and by far the most common way to optimize neural networks. At the same time, every state-of-the-art Deep Learning library contains implementations of various algorithms to optimize gradient descent (e.g. lasagne's\footnote{\url{http://lasagne.readthedocs.org/en/latest/modules/updates.html}}, caffe's\footnote{\url{http://caffe.berkeleyvision.org/tutorial/solver.html}}, and keras'\footnote{\url{http://keras.io/optimizers/}} documentation). These algorithms, however, are often used as black-box optimizers, as practical explanations of their strengths and weaknesses are hard to come by.

This article aims at providing the reader with intuitions with regard to the behaviour of different algorithms for optimizing gradient descent that will help her put them to use. In Section \ref{sec:gradient_descent_variants}, we are first going to look at the different variants of gradient descent. We will then briefly summarize challenges during training in Section \ref{sec:challenges}. Subsequently, in Section \ref{sec:algos}, we will introduce the most common optimization algorithms by showing their motivation to resolve these challenges and how this leads to the derivation of their update rules. Afterwards, in Section \ref{sec:parallelizing}, we will take a short look at algorithms and architectures to optimize gradient descent in a parallel and distributed setting. Finally, we will consider additional strategies that are helpful for optimizing gradient descent in Section \ref{sec:strategies}.

Gradient descent is a way to minimize an objective function $J(\theta)$ parameterized by a model's parameters $\theta \in \mathbb{R}^d$ by updating the parameters in the opposite direction of the gradient of the objective function $\nabla_\theta J(\theta)$ w.r.t. to the parameters. The learning rate $\eta$ determines the size of the steps we take to reach a (local) minimum. In other words, we follow the direction of the slope of the surface created by the objective function downhill until we reach a valley.\footnote{If you are unfamiliar with gradient descent, you can find a good introduction on optimizing neural networks at \url{http://cs231n.github.io/optimization-1/}.}

\section{Gradient descent variants} \label{sec:gradient_descent_variants}

There are three variants of gradient descent, which differ in how much data we use to compute the gradient of the objective function. Depending on the amount of data, we make a trade-off between the accuracy of the parameter update and the time it takes to perform an update.

\subsection{Batch gradient descent}

Vanilla gradient descent, aka batch gradient descent, computes the gradient of the cost function w.r.t. to the parameters $\theta$ for the entire training dataset:

\begin{equation}
\theta = \theta - \eta \cdot \nabla_\theta J( \theta)
\end{equation}

As we need to calculate the gradients for the whole dataset to perform just \emph{one} update, batch gradient descent can be very slow and is intractable for datasets that do not fit in memory. Batch gradient descent also does not allow us to update our model \emph{online}, i.e. with new examples on-the-fly.

In code, batch gradient descent looks something like this:

\begin{lstlisting}[language=python]
for i in range(nb_epochs):
  params_grad = evaluate_gradient(loss_function, data, params)
  params = params - learning_rate * params_grad
\end{lstlisting}

For a pre-defined number of epochs,  we first compute the gradient vector \lstinline{params_grad} of the loss function for the whole dataset w.r.t. our parameter vector \lstinline{params}.  Note that state-of-the-art deep learning libraries provide automatic differentiation that efficiently computes the gradient w.r.t. some parameters. If you derive the gradients yourself, then gradient checking is a good idea.\footnote{Refer to \url{http://cs231n.github.io/neural-networks-3/} for some great tips on how to check gradients properly.}

We then update our parameters in the direction of the gradients with the learning rate determining how big of an update we perform. Batch gradient descent is guaranteed to converge to the global minimum for convex error surfaces and to a local minimum for non-convex surfaces.

\subsection{Stochastic gradient descent}

Stochastic gradient descent (SGD) in contrast performs a parameter update for \emph{each} training example $x^{(i)}$ and label $y^{(i)}$:

\begin{equation}
\theta = \theta - \eta \cdot \nabla_\theta J( \theta; x^{(i)}; y^{(i)})
\end{equation}

\begin{figure}
	\centering
  	\includegraphics[width=0.4\linewidth]{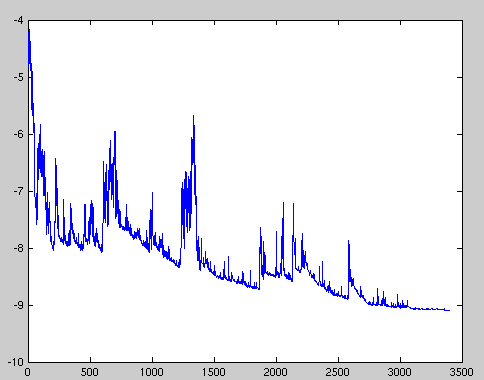}
  	\caption{SGD fluctuation (Source: \href{
https://upload.wikimedia.org/wikipedia/commons/f/f3/Stogra.png}{Wikipedia})}
  	\label{fig:sgd_fluctuation}
\end{figure}

Batch gradient descent performs redundant computations for large datasets, as it recomputes gradients for similar examples before each parameter update. SGD does away with this redundancy by performing one update at a time. It is therefore usually much faster and can also be used to learn online.
SGD performs frequent updates with a high variance that cause the objective function to fluctuate heavily as in Figure \ref{fig:sgd_fluctuation}.

While batch gradient descent converges to the minimum of the basin the parameters are placed in, SGD's fluctuation, on the one hand, enables it to jump to new and potentially better local minima. On the other hand, this ultimately complicates convergence to the exact minimum, as SGD will keep overshooting. However, it has been shown that when we slowly decrease the learning rate, SGD shows the same convergence behaviour as batch gradient descent, almost certainly converging to a local or the global minimum for non-convex and convex optimization respectively.
Its code fragment simply adds a loop over the training examples and evaluates the gradient w.r.t. each example. Note that we shuffle the training data at every epoch as explained in Section \ref{sec:shuffling}.

\begin{lstlisting}[language=python]
for i in range(nb_epochs):
  np.random.shuffle(data)
  for example in data:
    params_grad = evaluate_gradient(loss_function, example, params)
    params = params - learning_rate * params_grad
\end{lstlisting}

\subsection{Mini-batch gradient descent}

Mini-batch gradient descent finally takes the best of both worlds and performs an update for every mini-batch of $n$ training examples:

\begin{equation}
\theta = \theta - \eta \cdot \nabla_\theta J( \theta; x^{(i:i+n)}; y^{(i:i+n)})
\end{equation}

This way, it a) reduces the variance of the parameter updates, which can lead to more stable convergence; and b) can make use of highly optimized matrix optimizations common to state-of-the-art deep learning libraries that make computing the gradient w.r.t. a mini-batch very efficient. Common mini-batch sizes range between $50$ and $256$, but can vary for different applications. Mini-batch gradient descent is typically the algorithm of choice when training a neural network and the term SGD usually is employed also when mini-batches are used. Note: In modifications of SGD in the rest of this post, we leave out the parameters $x^{(i:i+n)}; y^{(i:i+n)}$ for simplicity.

In code, instead of iterating over examples, we now iterate over mini-batches of size $50$:

\begin{lstlisting}[language=python]
for i in range(nb_epochs):
  np.random.shuffle(data)
  for batch in get_batches(data, batch_size=50):
    params_grad = evaluate_gradient(loss_function, batch, params)
    params = params - learning_rate * params_grad
\end{lstlisting}

\section{Challenges} \label{sec:challenges}

Vanilla mini-batch gradient descent, however, does not guarantee good convergence, but offers a few challenges that need to be addressed:

\begin{itemize}
\item Choosing a proper learning rate can be difficult. A learning rate that is too small leads to painfully slow convergence, while a learning rate that is too large can hinder convergence and cause the loss function to fluctuate around the minimum or even to diverge.

\item Learning rate schedules \cite{Robbins1951} try to adjust the learning rate during training by e.g. annealing, i.e. reducing the learning rate according to a pre-defined schedule or when the change in objective between epochs falls below a threshold. These schedules and thresholds, however, have to be defined in advance and are thus unable to adapt to a dataset's characteristics \cite{Darken1992}.

\item Additionally, the same learning rate applies to all parameter updates. If our data is sparse and our features have very different frequencies, we might not want to update all of them to the same extent, but perform a larger update for rarely occurring features.

\item Another key challenge of minimizing highly non-convex error functions common for neural networks is avoiding getting trapped in their numerous suboptimal local minima. Dauphin et al. \cite{Dauphin2014} argue that the difficulty arises in fact not from local minima but from saddle points, i.e. points where one dimension slopes up and another slopes down. These saddle points are usually surrounded by a plateau of the same error, which makes it notoriously hard for SGD to escape, as the gradient is close to zero in all dimensions.
\end{itemize}

\section{Gradient descent optimization algorithms} \label{sec:algos}

In the following, we will outline some algorithms that are widely used by the Deep Learning community to deal with the aforementioned challenges. We will not discuss algorithms that are infeasible to compute in practice for high-dimensional data sets, e.g. second-order methods such as Newton's method\footnote{\url{https://en.wikipedia.org/wiki/Newton\%27s_method_in_optimization}}.

\subsection{Momentum}

\begin{figure}[!htb]
    \centering
    \begin{subfigure}{.5\textwidth}
        \centering
        \includegraphics[width=0.8\linewidth]{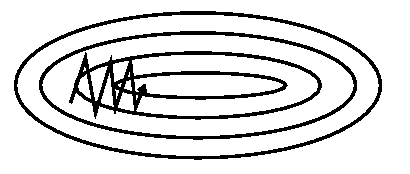}
        \caption{SGD without momentum}
        \label{fig:sgd_without_momentum}
    \end{subfigure}%
    \begin{subfigure}{0.5\textwidth}
        \centering
        \includegraphics[width=0.8\linewidth]{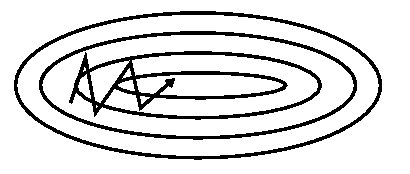}
        \caption{SGD with momentum}
        \label{fig:sgd_with_momentum}
    \end{subfigure}
    \caption{Source: \href{https://www.willamette.edu/~gorr/classes/cs449/momrate.html}{Genevieve B. Orr}}
\end{figure}

SGD has trouble navigating ravines, i.e. areas where the surface curves much more steeply in one dimension than in another \cite{Sutton1986}, which are common around local optima. In these scenarios, SGD oscillates across the slopes of the ravine while only making hesitant progress along the bottom towards the local optimum as in Figure \ref{fig:sgd_without_momentum}.

Momentum \cite{Qian1999} is a method that helps accelerate SGD in the relevant direction and dampens oscillations as can be seen in Figure \ref{fig:sgd_with_momentum}. It does this by adding a fraction $\gamma$ of the update vector of the past time step to the current update vector\footnote{Some implementations exchange the signs in the equations.}

\begin{align}
\begin{split}
v_t &= \gamma v_{t-1} + \eta \nabla_\theta J( \theta)\\
\theta &= \theta - v_t
\end{split}
\end{align}

The momentum term $\gamma$ is usually set to $0.9$ or a similar value.

Essentially, when using momentum, we push a ball down a hill. The ball accumulates momentum as it rolls downhill, becoming faster and faster on the way (until it reaches its terminal velocity, if there is air resistance, i.e. $\gamma < 1$). The same thing happens to our parameter updates: The momentum term increases for dimensions whose gradients point in the same directions and reduces updates for dimensions whose gradients change directions. As a result, we gain faster convergence and reduced oscillation.

\subsection{Nesterov accelerated gradient}

However, a ball that rolls down a hill, blindly following the slope, is highly unsatisfactory. We would like to have a smarter ball, a ball that has a notion of where it is going so that it knows to slow down before the hill slopes up again.

Nesterov accelerated gradient (NAG) \cite{Nesterov} is a way to give our momentum term this kind of prescience. We know that we will use our momentum term $\gamma \: v_{t-1}$ to move the parameters $\theta$. Computing $\theta - \gamma \: v_{t-1}$ thus gives us an approximation of the next position of the parameters (the gradient is missing for the full update), a rough idea where our parameters are going to be. We can now effectively look ahead by calculating the gradient not w.r.t. to our current parameters $\theta$ but w.r.t. the approximate future position of our parameters:

\begin{align}
\begin{split}
v_t &= \gamma \: v_{t-1} + \eta \nabla_\theta J( \theta - \gamma v_{t-1})\\
\theta &= \theta - v_t
\end{split}
\end{align}

\begin{figure}
	\centering
  	\includegraphics[width=0.4\linewidth]{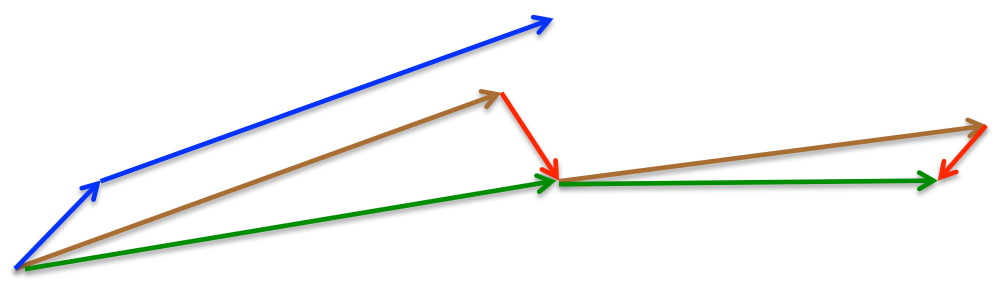}
  	\caption{Nesterov update (Source: \href{
http://www.cs.toronto.edu/~tijmen/csc321/slides/lecture_slides_lec6.pdf}{G. Hinton's lecture 6c})}
  	\label{fig:nesterov update}
\end{figure}

Again, we set the momentum term $\gamma$ to a value of around $0.9$. While Momentum first computes the current gradient (small blue vector in Figure \ref{fig:nesterov update}) and then takes a big jump in the direction of the updated accumulated gradient (big blue vector), NAG first makes a big jump in the direction of the previous accumulated gradient (brown vector), measures the gradient and then makes a correction (green vector). This anticipatory update prevents us from going too fast and results in increased responsiveness, which has significantly increased the performance of RNNs on a number of tasks \cite{Bengio2012a}.\footnote{Refer to \url{http://cs231n.github.io/neural-networks-3/} for another explanation of the intuitions behind NAG, while Ilya Sutskever gives a more detailed overview in his PhD thesis \cite{Sutskever2013a}.}

Now that we are able to adapt our updates to the slope of our error function and speed up SGD in turn, we would also like to adapt our updates to each individual parameter to perform larger or smaller updates depending on their importance.

\subsection{Adagrad}

Adagrad \cite{Duchi2011} is an algorithm for gradient-based optimization that does just this: It adapts the learning rate to the parameters, performing larger updates for infrequent and smaller updates for frequent parameters. For this reason, it is well-suited for dealing with sparse data. Dean et al. \cite{Dean2012} have found that Adagrad greatly improved the robustness of SGD and used it for training large-scale neural nets at Google, which -- among other things -- learned to recognize cats in Youtube videos\footnote{\url{http://www.wired.com/2012/06/google-x-neural-network/}}. Moreover, Pennington et al. \cite{Pennington2014} used Adagrad to train GloVe word embeddings, as infrequent words require much larger updates than frequent ones.

Previously, we performed an update for all parameters $\theta$ at once as every parameter $\theta_i$ used the same learning rate $\eta$. As Adagrad uses a different learning rate for every parameter $\theta_i$ at every time step $t$, we first show Adagrad's per-parameter update, which we then vectorize. For brevity, we set $g_{t, i}$ to be the gradient of the objective function w.r.t. to the parameter $\theta_i$ at time step $t$:

\begin{equation}
g_{t, i} = \nabla_{\theta_t} J( \theta_{t,i} )
\end{equation}

The SGD update for every parameter $\theta_i$ at each time step $t$ then becomes:

\begin{equation}
\theta_{t+1, i} = \theta_{t, i} - \eta \cdot g_{t, i}
\end{equation}

In its update rule, Adagrad modifies the general learning rate $\eta$ at each time step $t$ for every parameter $\theta_i$ based on the past gradients that have been computed for $\theta_i$:

\begin{equation}
\theta_{t+1, i} = \theta_{t, i} - \frac{\eta}{\sqrt{G_{t, ii} + \epsilon}} \cdot g_{t, i}
\end{equation}

$G_{t} \in \mathbb{R}^{d \times d} $ here is a diagonal matrix where each diagonal element $i, i$ is the sum of the squares of the gradients w.r.t. $\theta_i$ up to time step $t$\footnote{Duchi et al. \cite{Duchi2011} give this matrix as an alternative to the \emph{full} matrix containing the outer products of all previous gradients, as the computation of the matrix square root is infeasible even for a moderate number of parameters $d$.}, while $\epsilon$ is a smoothing term that avoids division by zero (usually on the order of $1e-8$). Interestingly, without the square root operation, the algorithm performs much worse.

As $G_{t}$ contains the sum of the squares of the past gradients w.r.t. to all parameters $\theta$ along its diagonal, we can now vectorize our implementation by performing an element-wise matrix-vector multiplication $\odot$ between $G_{t}$ and $g_{t}$:

\begin{equation}
\theta_{t+1} = \theta_{t} - \frac{\eta}{\sqrt{G_{t} + \epsilon}} \odot g_{t}.
\end{equation}

One of Adagrad's main benefits is that it eliminates the need to manually tune the learning rate. Most implementations use a default value of $0.01$ and leave it at that.

Adagrad's main weakness is its accumulation of the squared gradients in the denominator: Since every added term is positive, the accumulated sum keeps growing during training. This in turn causes the learning rate to shrink and eventually become infinitesimally small, at which point the algorithm is no longer able to acquire additional knowledge. The following algorithms aim to resolve this flaw.

\subsection{Adadelta}

Adadelta \cite{Zeiler2012} is an extension of Adagrad that seeks to reduce its aggressive, monotonically decreasing learning rate. Instead of accumulating all past squared gradients, Adadelta restricts the window of accumulated past gradients to some fixed size $w$.

Instead of inefficiently storing $w$ previous squared gradients, the sum of gradients is recursively defined as a decaying average of all past squared gradients. The running average $E[g^2]_t$ at time step $t$ then depends (as a fraction $\gamma $ similarly to the Momentum term) only on the previous average and the current gradient:

\begin{equation}
E[g^2]_t = \gamma E[g^2]_{t-1} + (1 - \gamma) g^2_t
\end{equation}

We set $\gamma$ to a similar value as the momentum term, around $0.9$. For clarity, we now rewrite our vanilla SGD update in terms of the parameter update vector $ \Delta \theta_t $:

\begin{align}
\begin{split}
\Delta \theta_t &= - \eta \cdot g_{t, i}\\
\theta_{t+1} &= \theta_t + \Delta \theta_t
\end{split}
\end{align}

The parameter update vector of Adagrad that we derived previously thus takes the form:

\begin{equation}
\Delta \theta_t = - \frac{\eta}{\sqrt{G_{t} + \epsilon}} \odot g_{t}
\end{equation}

We now simply replace the diagonal matrix $G_{t}$ with the decaying average over past squared gradients $E[g^2]_t$:

\begin{equation}
\Delta \theta_t = - \frac{\eta}{\sqrt{E[g^2]_t + \epsilon}} g_{t}
\end{equation}

As the denominator is just the root mean squared (RMS) error criterion of the gradient, we can replace it with the criterion short-hand:

\begin{equation}
\Delta \theta_t = - \frac{\eta}{RMS[g]_{t}} g_t
\end{equation}

The authors note that the units in this update (as well as in SGD, Momentum, or Adagrad) do not match, i.e. the update should have the same hypothetical units as the parameter. To realize this, they first define another exponentially decaying average, this time not of squared gradients but of squared parameter updates:

\begin{equation}
E[\Delta \theta^2]_t = \gamma E[\Delta \theta^2]_{t-1} + (1 - \gamma) \Delta \theta^2_t
\end{equation}

The root mean squared error of parameter updates is thus: 

\begin{equation}
RMS[\Delta \theta]_{t} = \sqrt{E[\Delta \theta^2]_t + \epsilon}
\end{equation}

Since $RMS[\Delta \theta]_{t}$ is unknown, we approximate it with the RMS of parameter updates until the previous time step. Replacing the learning rate $\eta $ in the previous update rule with $RMS[\Delta \theta]_{t-1}$ finally yields the Adadelta update rule:

\begin{align}
\begin{split}
\Delta \theta_t &= - \frac{RMS[\Delta \theta]_{t-1}}{RMS[g]_{t}} g_{t}\\
\theta_{t+1} &= \theta_t + \Delta \theta_t
\end{split}
\end{align}

With Adadelta, we do not even need to set a default learning rate, as it has been eliminated from the update rule.

\subsection{RMSprop}

RMSprop is an unpublished, adaptive learning rate method proposed by Geoff Hinton in Lecture 6e of his Coursera Class\footnote{\url{http://www.cs.toronto.edu/~tijmen/csc321/slides/lecture_slides_lec6.pdf}}.

RMSprop and Adadelta have both been developed independently around the same time stemming from the need to resolve Adagrad's radically diminishing learning rates. RMSprop in fact is identical to the first update vector of Adadelta that we derived above:

\begin{align}
\begin{split}
E[g^2]_t &= 0.9 E[g^2]_{t-1} + 0.1 g^2_t\\
\theta_{t+1} &= \theta_{t} - \frac{\eta}{\sqrt{E[g^2]_t + \epsilon}} g_{t}
\end{split}
\end{align}

RMSprop as well divides the learning rate by an exponentially decaying average of squared gradients. Hinton suggests $\gamma$ to be set to $0.9$, while a good default value for the learning rate $\eta$ is $0.001$.

\subsection{Adam}

Adaptive Moment Estimation (Adam) \cite{Kingma2015} is another method that computes adaptive learning rates for each parameter. In addition to storing an exponentially decaying average of past squared gradients $v_t$ like Adadelta and RMSprop, Adam also keeps an exponentially decaying average of past gradients $m_t$, similar to momentum:

\begin{align}
\begin{split}
m_t &= \beta_1 m_{t-1} + (1 - \beta_1) g_t\\
v_t &= \beta_2 v_{t-1} + (1 - \beta_2) g_t^2
\end{split}
\end{align}

$m_t$ and $v_t$ are estimates of the first moment (the mean) and the second moment (the uncentered variance) of the gradients respectively, hence the name of the method. As $m_t$ and $v_t$ are initialized as vectors of $0$'s, the authors of Adam observe that they are biased towards zero, especially during the initial time steps, and especially when the decay rates are small (i.e. $\beta_1$ and $\beta_2$ are close to $1$). 

They counteract these biases by computing bias-corrected first and second moment estimates:

\begin{align}
\begin{split}
\hat{m}_t &= \frac{m_t}{1 - \beta^t_1}\\
\hat{v}_t &= \frac{v_t}{1 - \beta^t_2}
\end{split}
\end{align}

They then use these to update the parameters just as we have seen in Adadelta and RMSprop, which yields the Adam update rule:

\begin{equation}
\theta_{t+1} = \theta_{t} - \frac{\eta}{\sqrt{\hat{v}_t} + \epsilon} \hat{m}_t
\end{equation}

The authors propose default values of $0.9$ for $\beta_1$, $0.999$ for $\beta_2$, and $10^{-8}$ for $\epsilon$. They show empirically that Adam works well in practice and compares favorably to other adaptive learning-method algorithms.

\subsection{AdaMax}

The $v_t$ factor in the Adam update rule scales the gradient inversely proportionally to the $\ell_2$ norm of the past gradients (via the $v_{t-1}$ term) and current gradient $|g_t|^2$:

\begin{equation}
v_t = \beta_2 v_{t-1} + (1 - \beta_2) |g_t|^2
\end{equation}

We can generalize this update to the $\ell_p$ norm. Note that Kingma and Ba also parameterize $\beta_2$ as $\beta^p_2$:

\begin{equation}
v_t = \beta_2^p v_{t-1} + (1 - \beta_2^p) |g_t|^p
\end{equation}

Norms for large $p$ values generally become numerically unstable, which is why $\ell_1$ and $\ell_2$ norms are most common in practice. However, $\ell_\infty$ also generally exhibits stable behavior. For this reason, the authors propose AdaMax \cite{Kingma2015} and show that $v_t$ with $\ell_\infty$ converges to the following more stable value. To avoid confusion with Adam, we use $u_t$ to denote the infinity norm-constrained $v_t$:

\begin{align}
\begin{split}
u_t &= \beta_2^\infty v_{t-1} + (1 - \beta_2^\infty) |g_t|^\infty\\
              & = \max(\beta_2 \cdot v_{t-1}, |g_t|)
\end{split}
\end{align}

We can now plug this into the Adam update equation by replacing $\sqrt{\hat{v}_t} + \epsilon$ with $u_t$ to obtain the AdaMax update rule:

\begin{equation}
\theta_{t+1} = \theta_{t} - \frac{\eta}{u_t} \hat{m}_t
\end{equation}

Note that as $u_t$ relies on the $\max$ operation, it is not as suggestible to bias towards zero as $m_t$ and $v_t$ in Adam, which is why we do not need to compute a bias correction for $u_t$. Good default values are again $\eta = 0.002$, $\beta_1 = 0.9$, and $\beta_2 = 0.999$.

\subsection{Nadam}

As we have seen before, Adam can be viewed as a combination of RMSprop and momentum: RMSprop contributes the exponentially decaying average of past squared gradients $v_t$, while momentum accounts for the exponentially decaying average of past gradients $m_t$. We have also seen that Nesterov accelerated gradient (NAG) is superior to vanilla momentum. 

Nadam (Nesterov-accelerated Adaptive Moment Estimation) \cite{Dozat2016a} thus combines Adam and NAG. In order to incorporate NAG into Adam, we need to modify its momentum term $m_t$. 

First, let us recall the momentum update rule using our current notation :

\begin{align}
\begin{split}
g_t &= \nabla_{\theta_t}J(\theta_t)\\
m_t &= \gamma m_{t-1} + \eta g_t\\
\theta_{t+1} &= \theta_t - m_t
\end{split}
\end{align}

where $J$ is our objective function, $\gamma$ is the momentum decay term, and $\eta$ is our step size. Expanding the third equation above yields:

\begin{equation} \label{eq:momentum}
\theta_{t+1} = \theta_t - ( \gamma m_{t-1} + \eta g_t)
\end{equation}

This demonstrates again that momentum involves taking a step in the direction of the previous momentum vector and a step in the direction of the current gradient.

NAG then allows us to perform a more accurate step in the gradient direction by updating the parameters with the momentum step \emph{before} computing the gradient. We thus only need to modify the gradient $g_t$ to arrive at NAG:

\begin{align}
\begin{split}
g_t &= \nabla_{\theta_t}J(\theta_t - \gamma m_{t-1})\\
m_t &= \gamma m_{t-1} + \eta g_t\\
\theta_{t+1} &= \theta_t - m_t
\end{split}
\end{align}

Dozat proposes to modify NAG the following way: Rather than applying the momentum step twice -- one time for updating the gradient $g_t$ and a second time for updating the parameters $\theta_{t+1}$ -- we now apply the look-ahead momentum vector directly to update the current parameters:

\begin{align} \label{eq:nag}
\begin{split}
g_t &= \nabla_{\theta_t}J(\theta_t)\\
m_t &= \gamma m_{t-1} + \eta g_t\\
\theta_{t+1} &= \theta_t - (\gamma m_t + \eta g_t)
\end{split}
\end{align}

Notice that rather than utilizing the previous momentum vector $m_{t-1}$ as in Equation \ref{eq:momentum}, we now use the current momentum vector $m_t$ to look ahead. In order to add Nesterov momentum to Adam, we can thus similarly replace the previous momentum vector with the current momentum vector. First, recall that the Adam update rule is the following (note that we do not need to modify $\hat{v}_t$):

\begin{align} 
\begin{split}
m_t &= \beta_1 m_{t-1} + (1 - \beta_1) g_t\\
\hat{m}_t & = \frac{m_t}{1 - \beta^t_1}\\
\theta_{t+1} &= \theta_{t} - \frac{\eta}{\sqrt{\hat{v}_t} + \epsilon} \hat{m}_t
\end{split}
\end{align}

Expanding the second equation with the definitions of $\hat{m}_t$ and $m_t$ in turn gives us:

\begin{equation}
\theta_{t+1} = \theta_{t} - \frac{\eta}{\sqrt{\hat{v}_t} + \epsilon} (\frac{\beta_1 m_{t-1}}{1 - \beta^t_1} + \frac{(1 - \beta_1) g_t}{1 - \beta^t_1})
\end{equation}

Note that $\frac{\beta_1 m_{t-1}}{1 - \beta^t_1}$ is just the bias-corrected estimate of the momentum vector of the previous time step. We can thus replace it with $\hat{m}_{t-1}$:

\begin{equation}
\theta_{t+1} = \theta_{t} - \frac{\eta}{\sqrt{\hat{v}_t} + \epsilon} (\beta_1 \hat{m}_{t-1} + \frac{(1 - \beta_1) g_t}{1 - \beta^t_1})
\end{equation}

This equation looks very similar to our expanded momentum term in Equation \ref{eq:momentum}. We can now add Nesterov momentum just as we did in Equation \ref{eq:nag} by simply replacing this bias-corrected estimate of the momentum vector of the previous time step $\hat{m}_{t-1}$ with the bias-corrected estimate of the current momentum vector $\hat{m}_t$, which gives us the Nadam update rule:

\begin{equation}
\theta_{t+1} = \theta_{t} - \frac{\eta}{\sqrt{\hat{v}_t} + \epsilon} (\beta_1 \hat{m}_t + \frac{(1 - \beta_1) g_t}{1 - \beta^t_1})
\end{equation}

\subsection{Visualization of algorithms}

The following two figures provide some intuitions towards the optimization behaviour of the presented optimization algorithms.\footnote{Also have a look at \url{http://cs231n.github.io/neural-networks-3/} for a description of the same images by Karpathy and another concise overview of the algorithms discussed.}

\begin{figure}[!htb]
    \centering
    \begin{subfigure}{.5\textwidth}
        \centering
        \includegraphics[width=0.8\linewidth]{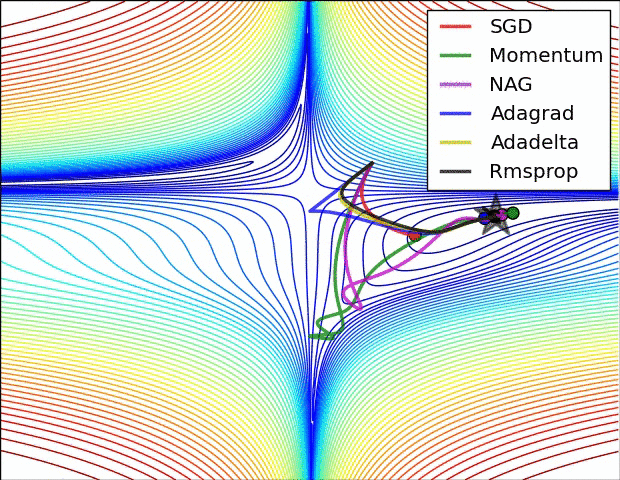}
        \caption{SGD optimization on loss surface contours}
        \label{fig:contours_evaluation_optimizers}
    \end{subfigure}%
    \begin{subfigure}{0.5\textwidth}
        \centering
        \includegraphics[width=0.8\linewidth]{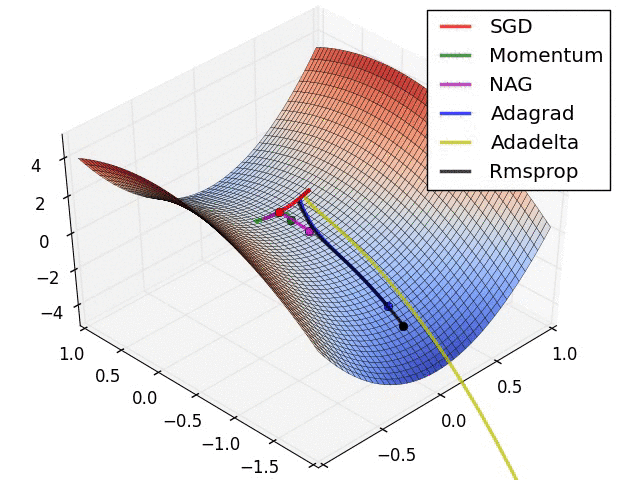}
        \caption{SGD optimization on saddle point}
        \label{fig:saddle_point_evaluation_optimizers}
    \end{subfigure}
    \caption{Source and full animations: \href{http://imgur.com/a/Hqolp}{Alec Radford}}
\end{figure}

In Figure \ref{fig:contours_evaluation_optimizers}, we see the path they took on the contours of a loss surface (the Beale function). All started at the same point and took different paths to reach the minimum. Note that Adagrad, Adadelta, and RMSprop headed off immediately in the right direction and converged similarly fast, while Momentum and NAG were led off-track, evoking the image of a ball rolling down the hill. NAG, however, was able to correct its course sooner due to its increased responsiveness by looking ahead and headed to the minimum.

Figure \ref{fig:saddle_point_evaluation_optimizers} shows the behaviour of the algorithms at a saddle point, i.e. a point where one dimension has a positive slope, while the other dimension has a negative slope, which pose a difficulty for SGD as we mentioned before. Notice here that SGD, Momentum, and NAG find it difficulty to break symmetry, although the latter two eventually manage to escape the saddle point, while Adagrad, RMSprop, and Adadelta quickly head down the negative slope, with Adadelta leading the charge.

As we can see, the adaptive learning-rate methods, i.e. Adagrad, Adadelta, RMSprop, and Adam are most suitable and provide the best convergence for these scenarios.

\subsection{Which optimizer to use?}

So, which optimizer should you use? If your input data is sparse, then you likely achieve the best results using one of the adaptive learning-rate methods. An additional benefit is that you will not need to tune the learning rate but will likely achieve the best results with the default value.

In summary, RMSprop is an extension of Adagrad that deals with its radically diminishing learning rates. It is identical to Adadelta, except that Adadelta uses the RMS of parameter updates in the numerator update rule. Adam, finally, adds bias-correction and momentum to RMSprop. Insofar, RMSprop, Adadelta, and Adam are very similar algorithms that do well in similar circumstances. Kingma et al. \cite{Kingma2015} show that its bias-correction helps Adam slightly outperform RMSprop towards the end of optimization as gradients become sparser. Insofar, Adam might be the best overall choice.

Interestingly, many recent papers use vanilla SGD without momentum and a simple learning rate annealing schedule. As has been shown, SGD usually achieves to find a minimum, but it might take significantly longer than with some of the optimizers, is much more reliant on a robust initialization and annealing schedule, and may get stuck in saddle points rather than local minima. Consequently, if you care about fast convergence and train a deep or complex neural network, you should choose one of the adaptive learning rate methods.

\section{Parallelizing and distributing SGD} \label{sec:parallelizing}

Given the ubiquity of large-scale data solutions and the availability of low-commodity clusters, distributing SGD to speed it up further is an obvious choice.
SGD by itself is inherently sequential: Step-by-step, we progress further towards the minimum. Running it provides good convergence but can be slow particularly on large datasets. In contrast, running SGD asynchronously is faster, but suboptimal communication between workers can lead to poor convergence. Additionally, we can also parallelize SGD on one machine without the need for a large computing cluster. The following are algorithms and architectures that have been proposed to optimize parallelized and distributed SGD.

\subsection{Hogwild!}

Niu et al. \cite{Niu2011} introduce an update scheme called Hogwild! that allows performing SGD updates in parallel on CPUs. Processors are allowed to access shared memory without locking the parameters. This only works if the input data is sparse, as each update will only modify a fraction of all parameters. They show that in this case, the update scheme achieves almost an optimal rate of convergence, as it is unlikely that processors will overwrite useful information.

\subsection{Downpour SGD}

Downpour SGD is an asynchronous variant of SGD that was used by Dean et al. \cite{Dean2012} in their DistBelief framework (the predecessor to TensorFlow) at Google. It runs multiple replicas of a model in parallel on subsets of the training data. These models send their updates to a parameter server, which is split across many machines. Each machine is responsible for storing and updating a fraction of the model's parameters. However, as replicas don't communicate with each other e.g. by sharing weights or updates, their parameters are continuously at risk of diverging, hindering convergence.

\subsection{Delay-tolerant Algorithms for SGD}

McMahan and Streeter \cite{Mcmahan2014} extend AdaGrad to the parallel setting by developing delay-tolerant algorithms that not only adapt to past gradients, but also to the update delays. This has been shown to work well in practice.

\subsection{TensorFlow}

TensorFlow\footnote{\url{https://www.tensorflow.org/}} \cite{Abadi2015a} is Google's recently open-sourced framework for the implementation and deployment of large-scale machine learning models. It is based on their experience with DistBelief and is already used internally to perform computations on a large range of mobile devices as well as on large-scale distributed systems. The distributed version, which was released in April 2016 \footnote{\url{http://googleresearch.blogspot.ie/2016/04/announcing-tensorflow-08-now-with.html}} relies on a computation graph that is split into a subgraph for every device, while communication takes place using Send/Receive node pairs.

\subsection{Elastic Averaging SGD}

Zhang et al. \cite{Zhang2014} propose Elastic Averaging SGD (EASGD), which links the parameters of the workers of asynchronous SGD with an elastic force, i.e. a center variable stored by the parameter server. This allows the local variables to fluctuate further from the center variable, which in theory allows for more exploration of the parameter space. They show empirically that this increased capacity for exploration leads to improved performance by finding new local optima.

\section{Additional strategies for optimizing SGD} \label{sec:strategies}

Finally, we introduce additional strategies that can be used alongside any of the previously mentioned algorithms to further improve the performance of SGD. For a great overview of some other common tricks, refer to \cite{LeCun1998}.

\subsection{Shuffling and Curriculum Learning} \label{sec:shuffling}

Generally, we want to avoid providing the training examples in a meaningful order to our model as this may bias the optimization algorithm. Consequently, it is often a good idea to shuffle the training data after every epoch. 

On the other hand, for some cases where we aim to solve progressively harder problems, supplying the training examples in a meaningful order may actually lead to improved performance and better convergence. The method for establishing this meaningful order is called Curriculum Learning \cite{Bengio2009a}. 

Zaremba and Sutskever \cite{Zaremba2014a} were only able to train LSTMs to evaluate simple programs using Curriculum Learning and show that a combined or mixed strategy is better than the naive one, which sorts examples by increasing difficulty.

\subsection{Batch normalization}

To facilitate learning, we typically normalize the initial values of our parameters by initializing them with zero mean and unit variance. As training progresses and we update parameters to different extents, we lose this normalization, which slows down training and amplifies changes as the network becomes deeper.

Batch normalization \cite{Ioffe2015a} reestablishes these normalizations for every mini-batch and changes are back-propagated through the operation as well. By making normalization part of the model architecture, we are able to use higher learning rates and pay less attention to the initialization parameters. Batch normalization additionally acts as a regularizer, reducing (and sometimes even eliminating) the need for Dropout.

\subsection{Early stopping}

According to Geoff Hinton: ``Early stopping (is) beautiful free lunch''\footnote{NIPS 2015 Tutorial slides, slide 63,  \url{http://www.iro.umontreal.ca/~bengioy/talks/DL-Tutorial-NIPS2015.pdf}}. You should thus always monitor error on a validation set during training and stop (with some patience) if your validation error does not improve enough.

\subsection{Gradient noise}

Neelakantan et al. \cite{Neelakantan2015} add noise that follows a Gaussian distribution $N(0, \sigma^2_t)$ to each gradient update:

\begin{equation}
g_{t, i} = g_{t, i} + N(0, \sigma^2_t)
\end{equation}

They anneal the variance according to the following schedule:
\begin{equation}
\sigma^2_t = \frac{\eta}{(1 + t)^\gamma}
\end{equation}

They show that adding this noise makes networks more robust to poor initialization and helps training particularly deep and complex networks. They suspect that the added noise gives the model more chances to escape and find new local minima, which are more frequent for deeper models.

\section{Conclusion}

In this article, we have initially looked at the three variants of gradient descent, among which mini-batch gradient descent is the most popular. We have then investigated algorithms that are most commonly used for optimizing SGD: Momentum, Nesterov accelerated gradient, Adagrad, Adadelta, RMSprop, Adam, AdaMax, Nadam, as well as different algorithms to optimize asynchronous SGD. Finally, we've considered other strategies to improve SGD such as shuffling and curriculum learning, batch normalization, and early stopping.

\bibliography{overview_sgd}
\bibliographystyle{plain}

\end{document}